\documentclass[runningheads]{llncs}
\usepackage{graphicx}
\usepackage{comment}
\usepackage{amsmath,amssymb} 
\DeclareMathOperator*{\argmax}{argmax}
\DeclareMathOperator*{\argmin}{argmin}
\usepackage{color}
\usepackage{parskip}
\usepackage[width=122mm,left=12mm,paperwidth=146mm,height=193mm,top=12mm,paperheight=217mm]{geometry}

\begin{document}

\pagestyle{headings}
\mainmatter
\def\CVPPPSubNumber{13}  
\def\ECCVSubNumber{\CVPPPSubNumber} 
\title{How useful is Active Learning for Image-based Plant Phenotyping?} 

\titlerunning{Active Learning for Image-based Plant Phenotyping}
%
\author{Koushik Nagasubramanian\ \and
Talukder Z. Jubery Author\ \and
Fateme Fotouhi Ardakani\ \and
Seyed Vahid Mirnezami \ \and
Asheesh K. Singh \ \and
Arti Singh \ \and
Soumik Sarkar \ \and Baskar Ganapathysubramanian }

\authorrunning{K.Nagasubramanian et al.}
%
\institute{Iowa State University, Ames IA 50011, USA 
\email{\{koushikn,znjubery,fotouhif,vahid,singhak,arti,soumiks,baskarg\}@iastate.edu}\\
}
\maketitle

\begin{abstract}
Deep learning models have been successfully deployed for a diverse array of image-based plant phenotyping applications including disease detection and classification. However, successful deployment of supervised deep learning models requires large amount of labeled data, which is a significant challenge in plant science (and most biological) domains due to the inherent complexity. Specifically, data annotation is costly, laborious, time consuming and needs domain expertise for phenotyping tasks, especially for diseases. To overcome this challenge, active learning algorithms have been proposed that reduce the amount of labeling needed by deep learning models to achieve good predictive performance.  Active learning methods adaptively select samples to annotate using an acquisition function to achieve maximum (classification) performance under a fixed labeling budget. We report the performance of four different active learning methods, (1) Deep Bayesian Active Learning (DBAL), (2) Entropy, (3) Least Confidence, and (4) Coreset, with conventional random sampling-based annotation for two different image-based classification datasets. The first image dataset consists of soybean [Glycine max L. (Merr.)] leaves belonging to eight different soybean stresses and a healthy class, and the second consists of nine different weed species from the field. For a fixed labeling budget, we observed that the classification performance of deep learning models with active learning-based acquisition strategies is better than random sampling-based acquisition for both datasets. The integration of active learning strategies for data annotation can help mitigate labelling challenges in the plant sciences applications particularly where deep domain knowledge is required.  
\end{abstract}



\section{Introduction and Related Work}
Deep learning architectures have advanced the state-of-the-art performance for image-based classification tasks \cite{Krizhevsky_Sutskever_Hinton_2012}, and have been successfully deployed for a diverse array of image-based plant phenotyping applications applications including disease detection, classification and quantification \cite{pound2017deep,singh2018deep}. However, one of the critical drawbacks of deep learning models is its necessity to have a large amount of labeled data to achieve good model accuracy. This is especially true for plant science applications, where annotating data can be costly, laborious, and time consuming to obtain, and generally need domain expertise (for instance, for plant stress image labeling that requires trained plant pathologists). To overcome this drawback, one effective and practical strategy is to use Active Learning (AL) based image annotation \cite{cohn1996active}. Weak supervision \cite{ghosal2019weakly}, domain adaptation \cite{valerio2019leaf,bellocchio2020combining}, domain randomization \cite{ward2020scalable}, synthetic dataset creation \cite{valerio2017arigan,cap2020leafgan,wu2020dcgan}, and transfer learning \cite{Tapas_2016} are some of  the other methods available to reduce the amount of labeling needed. However, when large amounts of unlabeled data is available but the task of labeling is hard or infeasible, AL methods are very useful. With the advent of high throughput phenotyping in plant sciences \cite{singh2016machine,Araus_Kefauver_Zaman-Allah_Olsen_Cairns_2018}, this dichotomy between increasingly large corpus of sensor and image data but our inability to exhaustively label them is expanding. AL methods adaptively select the most informative samples for labeling for the highest improvement in test accuracy. The goal of AL is to achieve maximum predictive performance under a fixed labeling budget, which makes it desirable for plant science applications.\par

\noindent Many AL methods have been proposed with different heuristics \cite{settles2009active} to reduce the amount of labeling needed for training machine learning (ML) models for classification tasks. A small amount of data is randomly chosen initially for labeling; and this labeled dataset is used to train a neural network model. Then, a batch of data from the remaining unlabeled data set is adaptively selected using an acquisition function for labeling by human domain experts. The acquisition function serves to select the most useful samples in the unlabeled dataset for improving neural network model performance. This process of choosing limited samples from unlabeled data sets, having the human expert annotate/label these limited samples, adding them to the labeled set, and retraining the model continues until one of two termination criteria is met --  a  desired performance threshold of the model is achieved, or the labeling budget is exhausted.\par 
\noindent Recently, in non-plant sciences problems, AL methods have been successfully applied for improving the performance of deep learning models, for example, deep learning based image classification \cite{Wang_Zhang_Li_Zhang_Lin_2016}, biomedical image segmentation \cite{Yang_Zhang_Chen_Zhang_Chen_2017}, text classification \cite{Yang_Zhang_Chen_Zhang_Chen_2017} and object detection \cite{Kao_Lee_Sen_Liu_2018}. In the field of plant phenotyping, uncertainty based sampling method was used to select samples for training Faster R-CNN model for panicle detection in cereal crops \cite{Chandra_Desai_Balasubramanian_Ninomiya_Guo_2020}. The continual improvement of AL strategies in the ML community, can be leveraged to significantly augment plant phenotyping efforts through state-of-the-art AL techniques. As a first step, there is a need to perform a comparative evaluation of the available sophisticated AL strategies in the context of canonical plant phenotyping applications. We compare four active learning methods defined by different acquisition functions: least confidence \cite{Culotta_McCallum_2005}, entropy \cite{Shannon_1948}, Deep Bayesian Active Learning \cite{Gal_Islam_Ghahramani_2017}, and core-set \cite{Sener_Savarese_2017} on two disparate plant phenotyping problems -- soybean stress identification \cite{Ghosal_Blystone_Singh_Ganapathysubramanian_Singh_Sarkar_2018} and weed species classification \cite{olsen2019deepweeds}.

\section{Materials and Methods}

\subsection{Datasets}
\subsubsection{Soybean Stress Dataset}
\begin{figure}
\centering
\includegraphics[height=6.5cm,width=12cm]{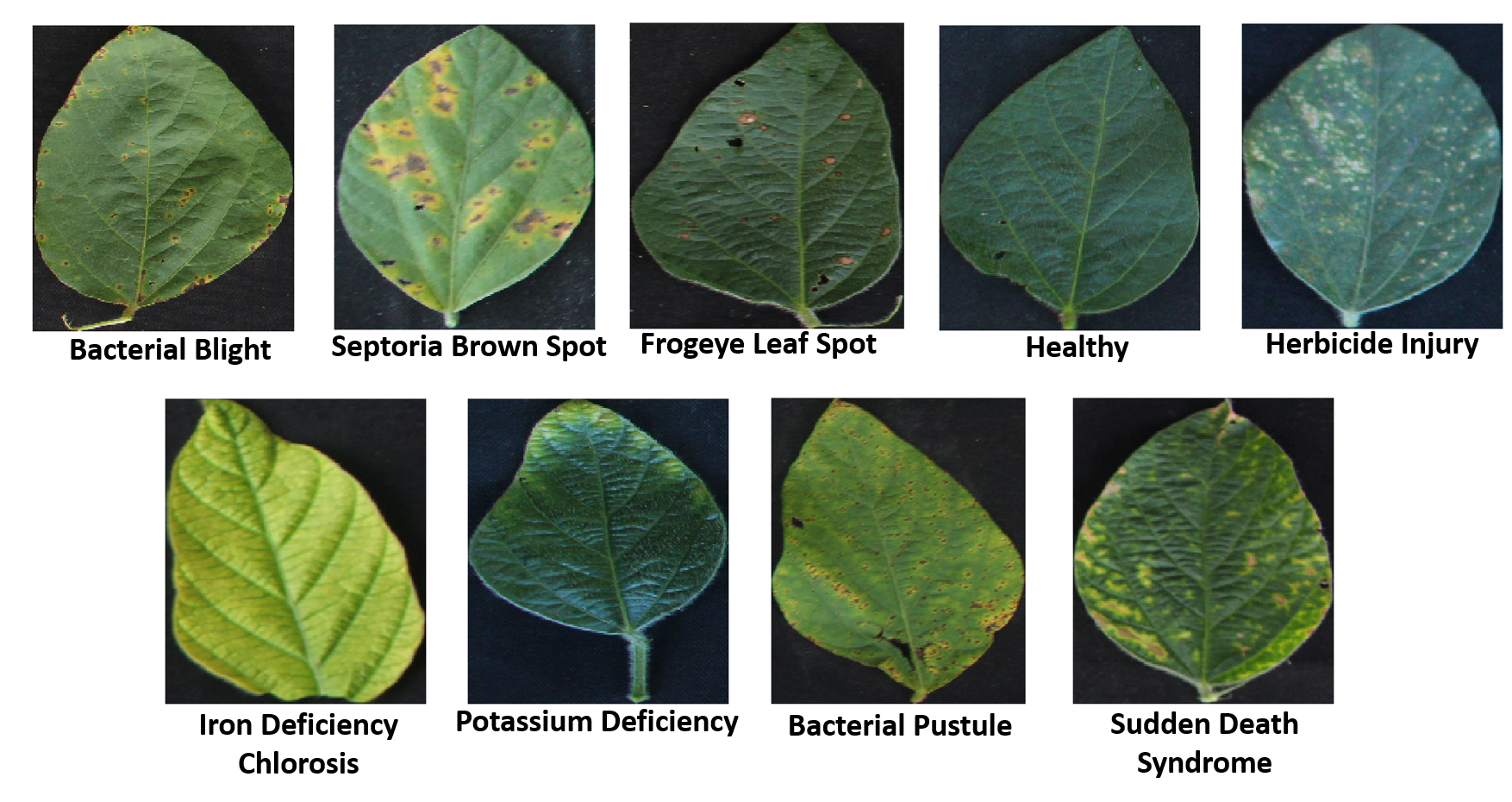}
\caption{The nine classes of data (eight stress, and one healthy) collected on soybean leaflets, which comprised the first data set}
\label{fig:soy_c}
\end{figure}

The dataset consists of 16,573 RGB images of soybean leaves across nine different classes (eight different soybean stresses, and the ninth class containing healthy soybean leaf). Details on the dataset can be found in \cite{Ghosal_Blystone_Singh_Ganapathysubramanian_Singh_Sarkar_2018}. Briefly, these classes cover a diverse spectrum of biotic and abiotic foliar stresses in soybean. Fig.~\ref{fig:soy_c} illustrates the nine different soybean leaf classes used in this study. The entire data set of 16573 images consisted of bacterial blight (No. of images = 1524), Septoria brown spot (= 1358), Frogeye leaf spot (= 1122), Healthy (= 4223), Herbicide injury (= 1395), Iron deficiency chlorosis (= 1844), Potassium deficiency (= 2186), bacterial pustule (= 1674), and sudden death syndrome (= 1247).
\subsubsection{Weed Species Dataset}
\begin{figure}[htb!]
\centering
\includegraphics[height=6.5cm,width=12cm]{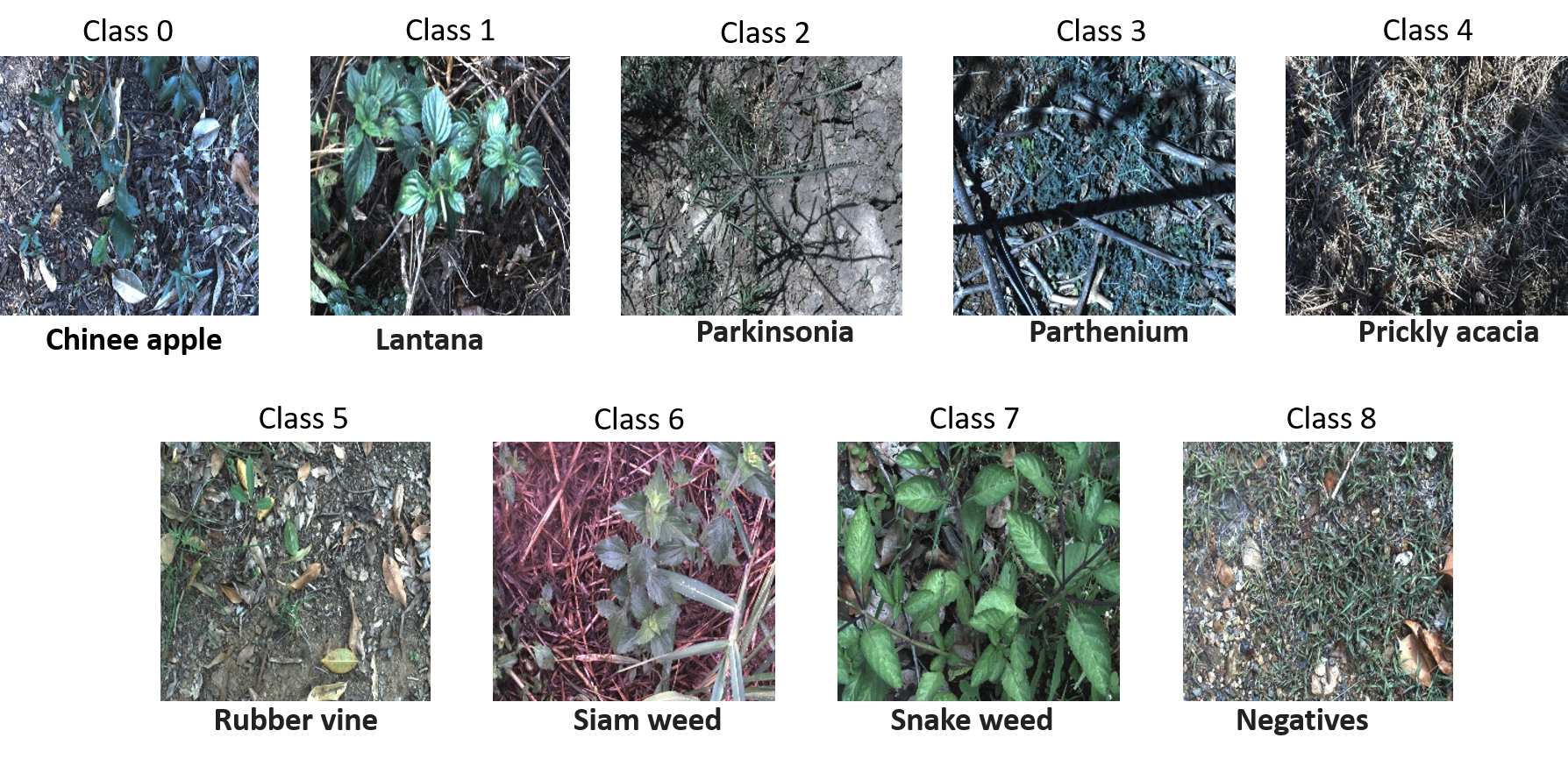}
\caption{The nine classes of data (eight stress, and one healthy) collected on soybean leaflets, which comprised the first data set}
\label{fig:weeds_c}
\end{figure}

The data set consists of 17,509 RGB images of weed species across nine different classes (eight weed classes and one non-weed class). Fig.~\ref{fig:weeds_c} illustrates the nine different classes used in this study, and the full description can be found in \cite{olsen2019deepweeds}. The entire data set of 17509 images consisted of Chinee apple (Ziziphus mauritiana) (No. of images = 1125), Lantana camara (= 1064), Parkinsonia aculeata (= 1031), Parthenium hysterophorus (= 1022), Prickly acacia (Acacia nilotica) (= 1062), Rubber vine (Cryptostegia grandiflora)  (= 1009), Siam weed (Chromolaena odorata) (= 1074), Snake weed (Stachytarpheta) (= 1016), and weed free (= 9106).
\subsection{Experimental Setup}
We trained a neural network-based classification model for identifying the class labels for input images in the two data sets, with the goal to achieve maximum classification performance for a fixed labeling budget. We evaluated each of the four active learning strategies (corset, DBAL, entropy, and least confidence) based on how well the neural network performed - i.e. using the classification accuracy on the complete dataset. We used MobileNetV2 \cite{Sandler_Howard_Zhu_Zhmoginov_Chen_2018} architecture for dataset \#1 (soybean stress classification) and ResNet-50 \cite{He_Zhang_Ren_Sun_2016} architecture for dataset \#2 (weed species classification). These networks were specifically chosen because of their popularity and strong performance, as well as to test the capability of AL on two distinct and well used networks. MobileNetV2 is a smaller, more compact network, while ResNet-50 is a large network, and were appropriate for dataset \#1 (controlled condition imaging) and dataset \#2 (field based imaging), respectively.\par
\begin{figure}
\centering
\includegraphics[height=6.5cm]{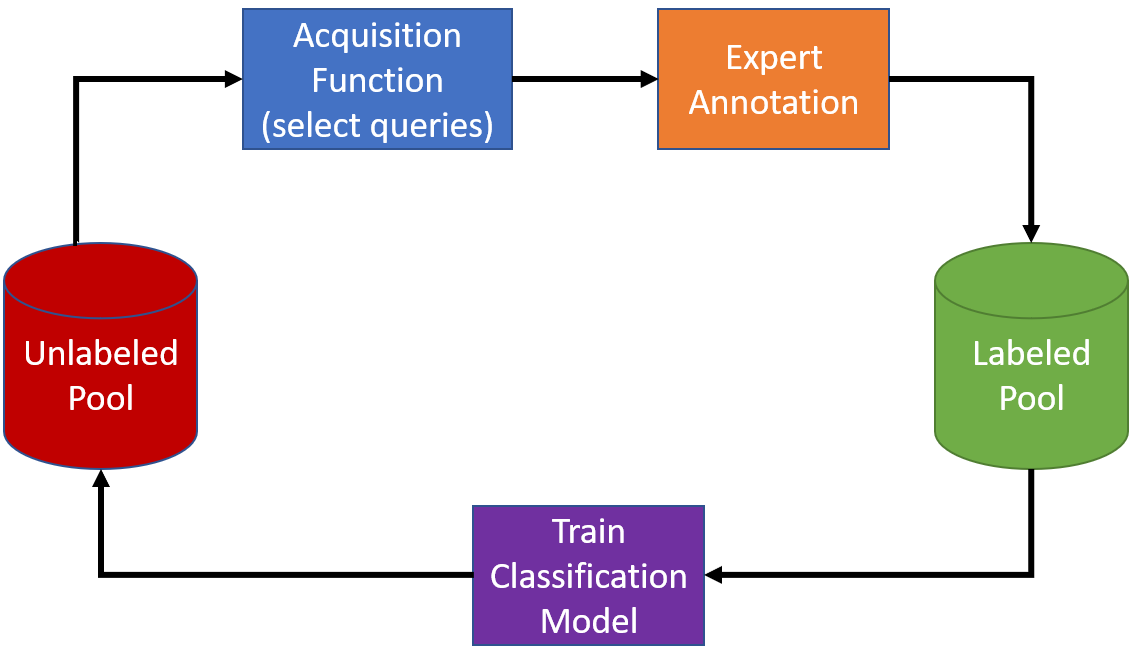}
\caption{Illustration of the pool-based active learning cycle for the soybean stress and weed stress classification datasets. The four method of active learning included: Least Confidence, Entropy, Deep Bayesian Active Learning, and Core-set.}
\label{fig:flow_c}
\end{figure}

 
\noindent Each data set was analyzed separately. The AL approach was repeated ten times for each dataset. For each run, we randomly selected and labeled 5$\%$ of the samples from the complete data to create a validation set . Each run starts with an initial random batch of 1000 samples spread across different classes, which was used for the evaluation of all four active learning methods. After training the neural network model for 100 epochs, a query batch of 1000 samples from the remaining unlabeled dataset was selected. This selection was performed using the acquisition function of each of the four active learning algorithms (so each AL approach will potentially select distinct set of 1000 samples to next annotate).  These 1000 samples were added to the labeled dataset, to retrain the neural network model. This process was repeated until the labeling budget was exhausted (labeling budget was 10,000 samples for the soybean stress classification, and 9000 samples for the weed species classification). We saved the model with best validation accuracy. The model was retrained from scratch after every selection of new labeled samples for 100 epochs. We optimized the model using Adam \cite{Kingma_Ba_2014}  optimizer with the default learning rate of 0.001. We used Keras \cite{Chollet_others_2015} with Tensorflow \cite{abadi2016tensorflow} backend for the implementation. A schematic of the approach is shown in Fig.~\ref{fig:flow_c}.
\subsection{Evaluated Methods}
Formally, let $x$ be the input and $y$ $\epsilon$ $(1, \dots, N)$ be the output of the classification model in the active learning setup.
The neural network was trained using labelled set $\boldsymbol{L_{pool}}$. 
The active learning methods selects a batch of \textbf{b} points [$x_1^*,\dots,x_b^* $]  from the unlabeled pool $\boldsymbol{U_{pool}}$ for expert annotation according to an acquisition criterion, and add these \textbf{b} points to the labeled set $\boldsymbol{L_{pool}}$. The four active learning methods are described below:
\subsubsection{Random.}
 The samples are chosen at random from the
 unlabeled data. This represents the baseline, if AL methods are not used. 
 \subsubsection{Least Confidence.}
 The unlabeled samples are sorted in
 ascending order according to maximum predicted classification probability for the input $x$ $(p(\frac{y}{x}))$ and the samples with the lower rank are chosen for labeling \cite{Culotta_McCallum_2005}. 
 \begin{align}
     [x_1^*,\dots,x_b^*]=\argmin_{[x_1^*,\dots,x_b^*]\subseteq\boldsymbol{U_{pool}}} \max_{k=1,\dots,N}p\left(\frac{y=k}{x}\right)
 \end{align}
\subsubsection{Entropy.}The unlabeled samples with highest entropy H of the predicted classification probability distribution p are chosen for labeling \cite{Shannon_1948}.
\begin{align}
    H\left( \frac{y}{x},\boldsymbol{L_{pool}} \right)=-\sum_{k=1}^{N}p\left(\frac{y=k}{x}\right)log\left(\frac{y=k}{x}\right)
\end{align}

\begin{align}
    [x_1^*,\dots,x_b^*]=\argmax_{[x_1^*,\dots,x_b^*]\subseteq\boldsymbol{U_{pool}}} H\left( \frac{y}{x},\boldsymbol{L_{pool}}\right)
\end{align}
\subsubsection{Core-set.}A set of diverse samples that best represents the distribution of the entire dataset in the representation space (penultimate layer) learned by the neural network model are chosen for labeling. The greedy approximation method was used to implement the core-set selection \cite{Sener_Savarese_2017}.
\subsubsection{Deep Bayesian Active Learning (DBAL).}
The Monte Carlo dropout (MC-dropout) \cite{Gal_Ghahramani_2016} based uncertainty estimation is combined  with Bayesian Active Learning by Disagreement (BALD) \cite{houlsby2011bayesian} acquisition framework for selecting the samples in DBAL \cite{Gal_Islam_Ghahramani_2017}. The MC-dropout based uncertainty estimates were computed by averaging the outputs of T different forward stochastic passes through the trained neural network model with weights $w_t$ for the pass $t$ during the test time. A new dropout mask was applied during each of the $T$ forward passes. The BALD acquisition function calculates the mutual information between the data samples and the model weights. Unlabeled data samples with larger mutual information between the predicted label and model weights were selected for labeling. The uncertainty estimate $p$ is:
 \begin{align}
    p\left( \frac{y}{x},\boldsymbol{L_{pool}} \right)=\frac{1}{T}\sum_{t=1}^{T}p\left(\left(\frac{y=k}{x}\right),w_t\right)
 \end{align}
 The BALD acquisition criterion I is:
 \begin{equation}
 \begin{split}
    I\left( \frac{y}{x},\boldsymbol{L_{pool}} \right)=H\left( \frac{y}{x},\boldsymbol{L_{pool}} \right)\\- \frac{1}{T}\sum_{t=1}^{T}\sum_{k=1}^{N}p\left(\left(\frac{y=k}{x}\right),w_t\right)log\left(\left(\frac{y=k}{x}\right),w_t\right)
 \end{split}
 \end{equation}
 \begin{align}
      [x_1^*,\dots,x_b^*]=\argmax_{[x_1^*,\dots,x_b^*]\subseteq\boldsymbol{U_{pool}}}I\left( \frac{y}{x},\boldsymbol{L_{pool}} \right)
 \end{align}
 \section{Results and Discussion}
\begin{figure}
\centering
\includegraphics[height=6.5cm]{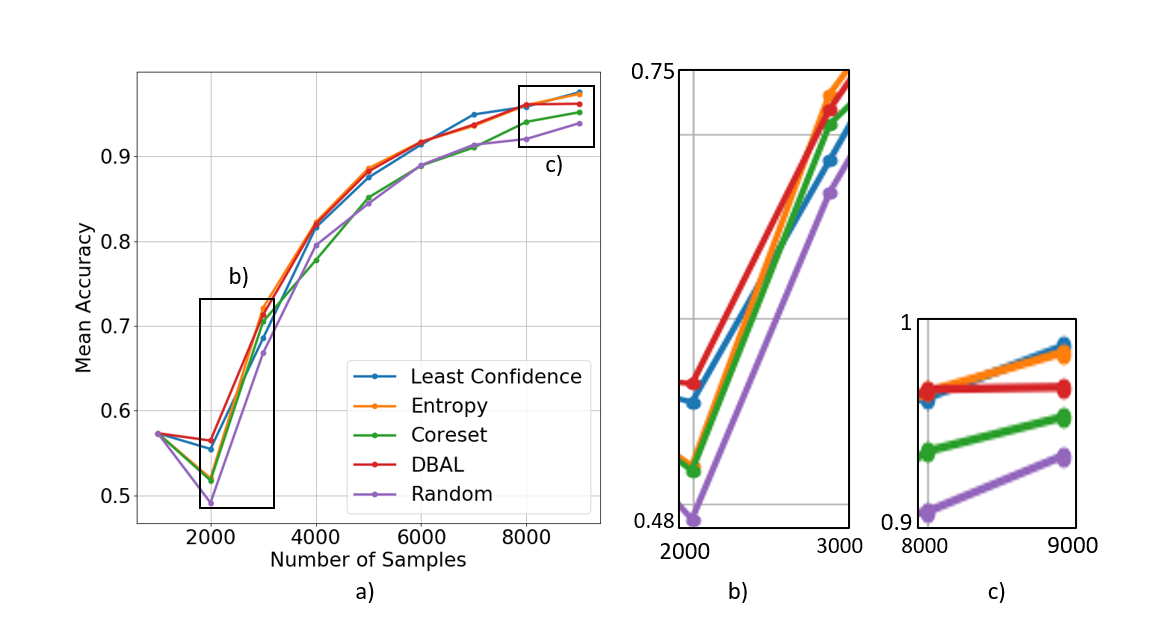}
\caption{a) MobileNetV2 accuracy plots of different active learning algorithms for soybean stress classification dataset. The results were averaged over 10 experiments. We show the zoomed in image of the accuracy plot for some points in b) and c).}
\label{fig:soy_z}
\end{figure}

\begin{figure}
\centering
\includegraphics[height=6.5cm]{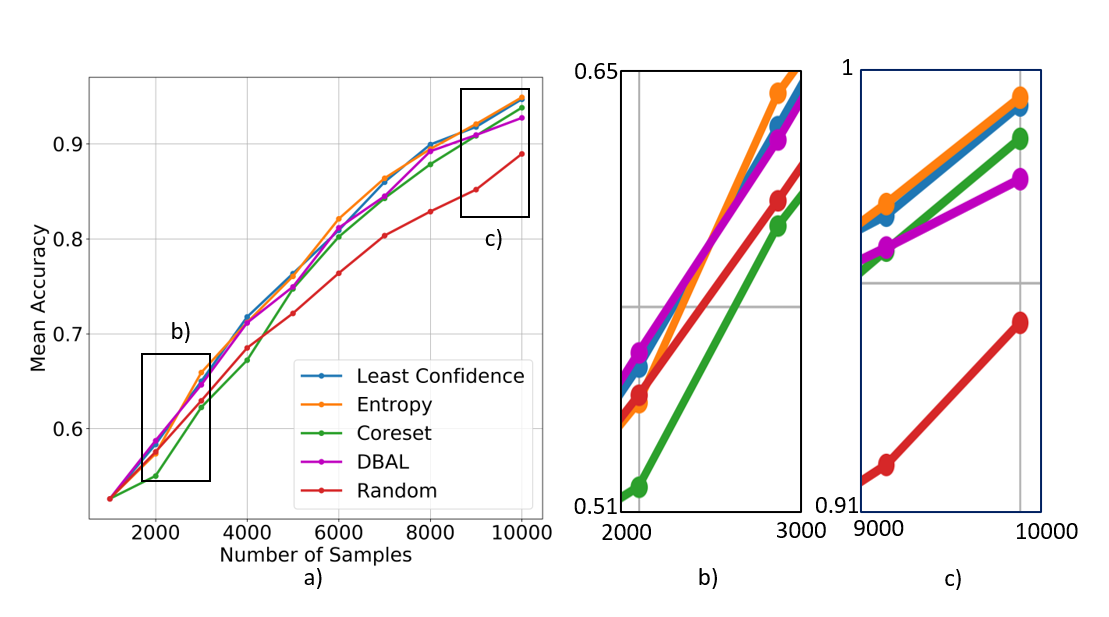}
\caption{a) ResNet50 accuracy plots of different active learning algorithms for weed species classification dataset. The results were averaged over 10 experiments. We show the zoomed in image of the accuracy plot for some points in b) and c).}
\label{fig:weeds_z}
\end{figure}

 Mean accuracy for different active learning methods for the two canonical problems on soybean stress classification and weed species classification are presented in Fig.~\ref{fig:soy_z}. and Fig.~\ref{fig:weeds_z} respectively.
For the soybean stress classification dataset, we clearly observe that all the uncertainty sampling based active learning methods outperform random sampling whereas diversity sampling based Coreset method underperforms. For the weed species classification dataset, all active learning algorithms outperform random sampling.  The performance gain due to AL methods over random sampling for plant domain datasets is similar to the improvement observed in other domain datasets like MNIST and CIFAR10 \cite{beluch2018power}. The overall performance gains of active learning algorithms were higher for weed species dataset than soybean stress dataset. One reason for this could be the challenging nature of the weed species dataset which was collected under diverse field conditions whereas the soybean dataset was collected under indoor conditions with constant illumination. Additionally, the field images for weed data set had more background objects and obscurity compared to the soybean dataset, which was images under more controlled conditions \cite{Ghosal_Blystone_Singh_Ganapathysubramanian_Singh_Sarkar_2018}. Hence, the random sampling-based annotation method provides a stronger baseline for the soybean stress dataset. The dip in accuracy at 2000 samples for the soybean dataset was due to high class-imbalance in the labeled dataset after the sample selection.\par
\begin{figure}
\centering
\includegraphics[height=10cm]{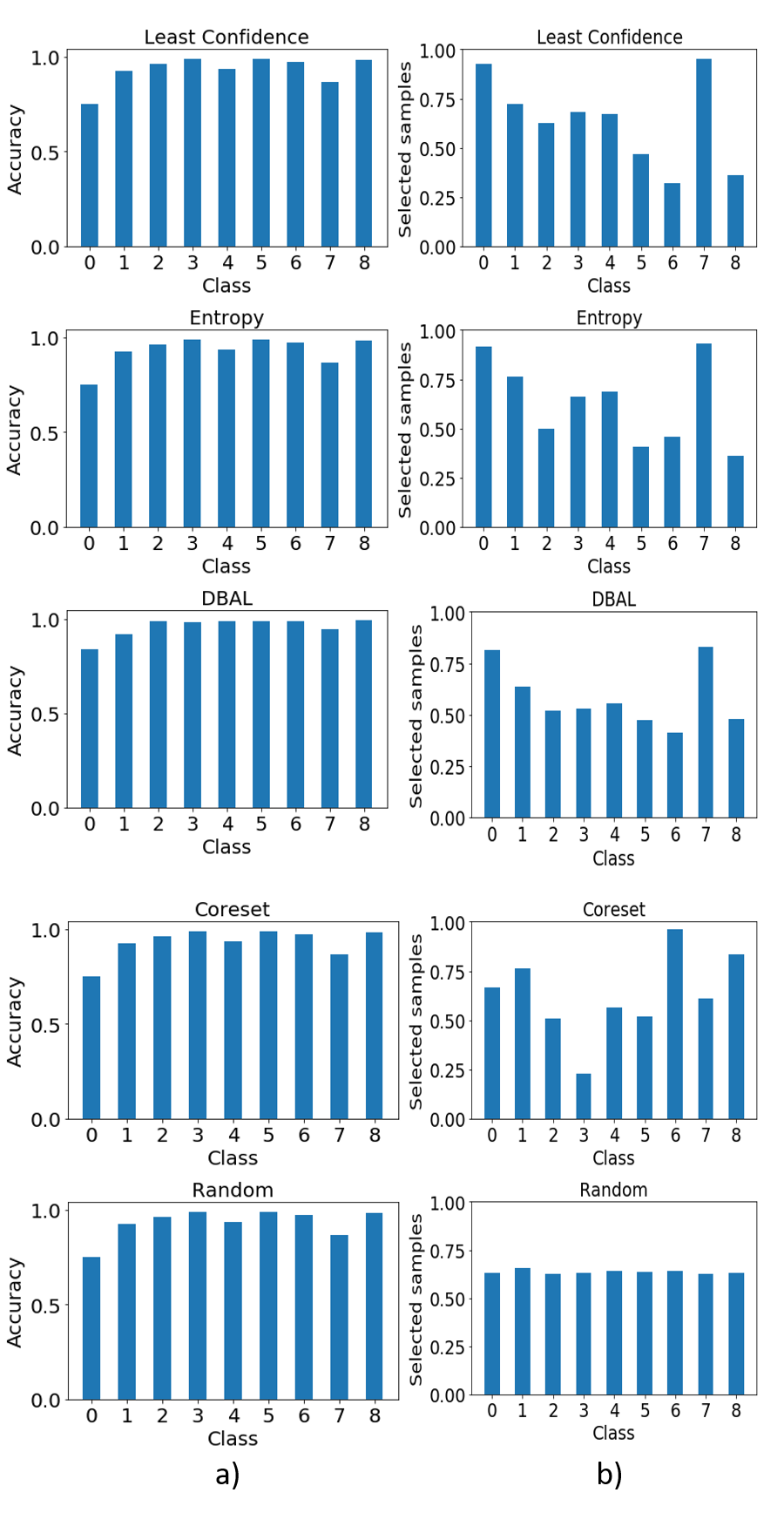}
\caption{a) An example of per class classification accuracy of MobileNetV2 model on soybean stress classification dataset using different active learning algorithms from a single experiment for a labeling budget of 9000 samples. b) Per class sample selection percentage (Number of sample selected from a class/Total number of samples available in a class) of different active learning algorithms for the results shown in a). The nine classes are as following: 0 = bacterial blight, 1 = Septoria Brown Spot, 2 = Frogeye Leaf Spot, 3= Healthy, 4= Herbicide Injury, 5 = Iron Deficiency Chlorosis, 6 = Potassium Deficiency, 7 = Bacterial Pustule, 8 = Sudden Death Syndrome.}
\label{fig:soy_sel}
\end{figure}
\begin{figure}
\centering
\includegraphics[height=10cm]{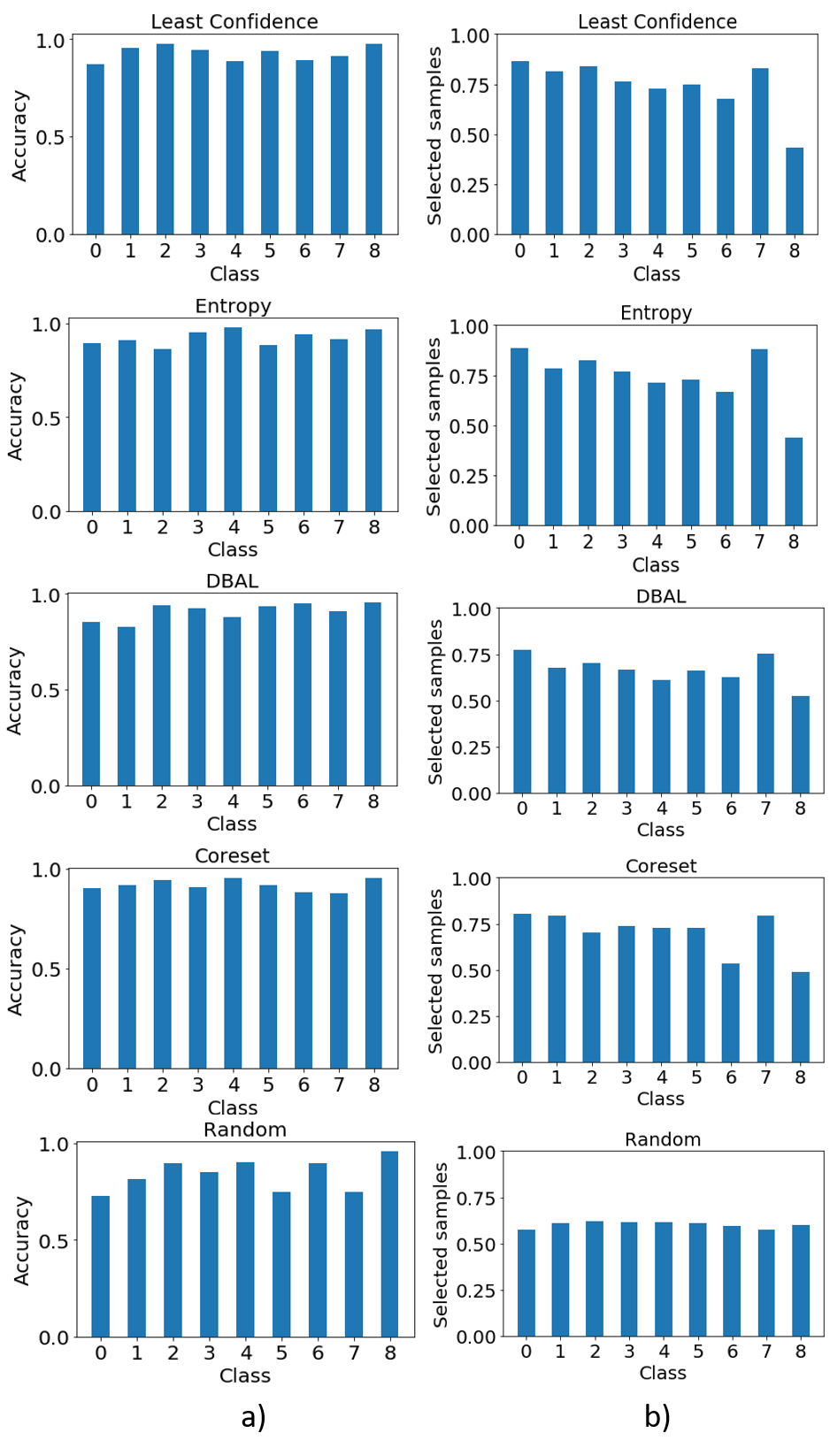}
\caption{a) An example of per class classification accuracy of MobileNetV2 model on soybean stress classification dataset using different active learning algorithms from a single experiment for a labeling budget of 9000 samples. b) Per class sample selection percentage (Number of sample selected from a class/Total number of samples available in a class) of different active learning algorithms for the results shown in a). The nine classes are as following: 0 = bacterial blight, 1 = Septoria Brown Spot, 2 = Frogeye Leaf Spot, 3= Healthy, 4= Herbicide Injury, 5 = Iron Deficiency Chlorosis, 6 = Potassium Deficiency, 7 = Bacterial Pustule, 8 = Sudden Death Syndrome.}
\label{fig:weeds_sel}
\end{figure}

\noindent \underline{\textbf{Which samples are selected?}}: A random selection strategy is expected to blindly pick new samples for annotation, therefore the distribution of selected points is expected to be uniform. In contrast, we anticipate the AL based methods to pick fewer samples from classes that are well predicted, and instead pick more points from classes that are not well predicted, for example class ‘0’ and ‘7’ are both bacterial diseases with very similar and confounding symptoms, which are at times even difficult for human raters to distinguish. We visualize this expected behavior in Fig.~\ref{fig:soy_sel} (for soybean stress classification) and Fig.~\ref{fig:weeds_sel} (for weed classification). The per-class classification accuracy of different active learning methods and random sampling is shown in Fig.~\ref{fig:soy_sel}a and Fig.~\ref{fig:weeds_sel}a respectively. The per class sample selection percentage, i.e. how many samples are used per class (calculated as number of sample selected from a class/total number of samples available in a class) of different active learning algorithms are presented in Fig.~\ref{fig:soy_sel}b and Fig.~\ref{fig:weeds_sel}b respectively. The accuracy plot of random method indicates the classes that are hard and easy to predict. The clear inverse relationship between the performance of individual classes shown in the accuracy plot of random and the number of samples chosen is apparent across all uncertainty based AL methods. This is in stark contrast to a naïve random sampling (Figure of Fig.~\ref{fig:soy_sel}b and Fig.~\ref{fig:weeds_sel}b). 
Class ‘0’ and Class ‘7’ have low per-class classification accuracy from random sampling-based annotation (Fig.~\ref{fig:soy_sel}). Least Confidence, Entropy and DBAL methods chose more samples from the Class 0 and Class 7 and obtained better per-class accuracy than random sampling. AL based like LC, Entropy and DBAL methods selected more samples from stresses which are highly confusing when compared to less confusing stresses. This is very promising from a domain perspective because confounding symptoms classes are more extensively sampled by these three AL methods. The uncertainty-based methods sampled only a small percentage of samples from classes that have high per-class accuracy for random sampling (Classes 3, 5, 6 and 8) method. In contrast to the uncertainty based acquisition functions of LC, Entropy, and DBAL, core-set uses a diversity based sampling. Its comparatively poor performance can be explained by the fact that it chooses less samples from classes exhibiting less diversity, even if that class is difficult to classify. 
Uncertainty based AL algorithms adaptively sampled more from the low accuracy classes of random sampling method (Classes 0, 1, and 7) and sampled less from the  high accuracy classes of random sampling methods (Classes 6 and 8). There was no consistent trend for leaf variation (narrow and broad leaved).The AL methods show promising results for plant sciences problems where extensive data are needed to train useable models. These include diverse applications, such as cluttered image problem for soybean cyst nematode egg detection \cite{akintayo2018deep}, hyperspectral imaging \cite{nagasubramanian2018explaining,roscher2016detection}, abiotic stress disease rating \cite{naik2017real,zhang2017computer}, root imaging \cite{falk2020computer,falk2020soybean}, yield predictors \cite{parmley2019development,parmley2019machine} et cetera.  

\section{Conclusions}
In this paper, we compared four different active learning algorithms for classification of soybean stress and weed species datasets. We empirically showed that uncertainty based active learning methods outperform random sampling-based annotation for classification of two distinct datasets using MobileNetV2 and ResNet50 architectures. Uncertainty based active learning strategy reasonably outperformed random sampling, especially in the lower end of the labeled dataset availability. For the two cases shown here, Entropy sampling is marginally better than the other AL strategies. We believe that active learning methods can be quite helpful in reducing the amount of labeling needed for image-based plant phenotyping tasks. In future, AL methods should be combined with unsupervised representation learning methods to further increase the label efficiency. 
\section{Acknowledgements}
We thank Olsen et al., 2019 for making the weed dataset publicly available. We thank J. Brungardt, B. Scott, H. Zhang, and undergraduate students (A.K.S. laboratory) for help in imaging and data collection; A. Lofquist and J. Stimes (B.G. laboratory) for developing the marking app and data storage back-end.
\section{Funding Information}
This work was funded by the Iowa Soybean Association (A.S.), an Iowa State University (ISU) internal grant (to all authors), a NSF/USDA National Institute of Food and Agriculture grant 2017-67007-26151 (to all authors), a Monsanto Chair in Soybean Breeding at Iowa State University (A.K.S.), Raymond F. Baker Center for Plant Breeding at Iowa State University (A.K.S.), an ISU Plant Science Institute fellowship (to B.G., A.K.S., and S.S.), and USDA IOW04403 (to A.S. and A.K.S.).
\section{Code Availability}
All the codes for the active learning approaches described in this work are available for the community at https://github.com/koushik-n/Active-Learning-Plant-Phenotyping.
\section{Author Contributions}
KN, TJ, AS, AKS, SS, BG formulated the problem and designed the study; KN, TJ, SM, FF developed software and ran numerical experiments;  KN, FF analyzed data with help from all authors; All authors contributed to writing the paper. 
\section{Conflict of Interest}
The authors declare that they have no conflicts of interest.
\bibliographystyle{splncs04}
\bibliography{Al_ref.bib}
\end{document}